\documentclass[11pt,a4paper]{article}
\usepackage[hyperref]{emnlp2018}
\usepackage{times}
\usepackage{latexsym}

\usepackage{url}

\usepackage{amsmath,amssymb,amsfonts}
\usepackage{algorithmic}
\usepackage{graphicx}
\usepackage{textcomp}
\usepackage{xcolor}

\usepackage{adjustbox}
\usepackage{diagbox}
\usepackage{array}

\usepackage{colortbl}

\usepackage{makecell}

\newcolumntype{R}[2]{%
    >{\adjustbox{angle=#1,lap=\width-(#2)}\bgroup}%
    r%
    <{\egroup}%
}

\usepackage{makecell}

\usepackage{multirow}
\usepackage{hhline}
\usepackage{tabularx}

\aclfinalcopy 

\title{Fast Approach to Build an Automatic Sentiment Annotator for Legal Domain using Transfer Learning}

 \author{Viraj Gamage, Menuka Warushavithana, Nisansa de Silva, \\ {\bf Amal Shehan Perera}, {\bf Gathika Ratnayaka} \and {\bf Thejan Rupasinghe}  \\
Department of Computer Science \& Engineering \\ University of Moratuwa \\
  {\tt viraj.14@cse.mrt.ac.lk}}

\date{}

\begin{document}
\maketitle
\begin{abstract}
This study proposes a novel way of identifying the sentiment of the phrases used in the legal domain. The added complexity of the language used in law, and the inability of the existing systems to accurately predict the sentiments of words in law are the main motivations behind this study. This is a transfer learning approach, which can be used for other domain adaptation tasks as well. The proposed methodology achieves an improvement of over 6\% compared to the source model's accuracy in the legal domain.
\end{abstract}

\section{Introduction}
Sentiment analysis tasks are a common component in many Natural Language Processing (NLP) applications. As described by~\citet{esuli2007sentiwordnet}, sentiment analysis or \textit{sentiment classification} is a recent methodology that aligns with information retrieval and computational linguistics which is focused on the opinion towards something which is represented by a certain text.  

In many recent studies involving NLP in various domains, it is common to reuse the seminal RNTN (Recursive Neural Tensor Network) model~\cite{socher2013recursive} trained on movie reviews for sentiment analysis. However, it is obvious that this trained model has bias towards the movie review text on which it is based. The traditional way to remedy this problem is to retrain the model from scratch using the same algorithm. But, the algorithm proposed by~\citet{socher2013recursive} is quite manual labour intensive given that it requires a significantly large enough corpus annotated on sentiment manually. This difficulty is the reason for most natural language processing applications to reuse the original model despite the mismatch between the trained domain and the domain to which it is being applied.      

Law is a field involving grand collisions of ideas, most of which are in the form of written text, thus open to linguistic research. However, the language used in these documents is rather complex and esoteric to a certain degree, which makes it challenging to be utilized in intelligent systems. Lawyers, paralegals, and other legal professionals spend a considerable part of their time reading transcripts of past court cases, taking notes and collecting precedents to make their case stronger in court. This task is cumbersome and time-consuming. Therefore, it is an open opportunity for computer scientists to introduce efficient methods and tools for the legal domain. From this point onward we shall be referring to the \textit{court case transcripts} as \textit{court cases}. 

In this study, we propose a novel way of applying sentiment analysis on the contents (words/phrases) of court cases. This analysis is useful in the NLP or Natural Language Understanding (NLU) tasks where it is vital to identify the stakeholder bias in each of the statements. Similarly, sentiment analysis in legal text can become useful in automating the following tasks related to legal literature.

\begin{itemize}
    \item Identifying the arguments in a court case
    \item Identifying the arguments which were supportive or against for a certain party in a court case
    \item Identifying or synthesizing counter-arguments for a given argument in a court case
\end{itemize}   

To identify the application of sentiment classification in the legal domain, consider the following example which was extracted from a legal case~\cite{2017lee}.\\

\noindent\textit{The District Court concluded that Lee's counsel had performed deficiently.}\\

In the above example, the phrase \textit{had performed deficiently}  induces a negative sentiment towards Lee's counsel. The sentiment of \textit{concluded that} denotes that the court agrees with the inner sentence. Complete sentence denotes that court's opinion towards Lee's counsel is negative. Consider the following extracted from the same case,\\

\noindent\textit{...the Government conceded that Lee's counsel had performed deficiently.}\\

This sentence contains the same inner sentence but in the legal domain the phrase called \textit{conceded that} indicates a situation where the government initially disagreed but eventually had to agree. That phrase induces a negative sentiment on the inner sentence which is negative towards Lee's counsel. Therefore, it is fair to assume that the government and Lee's counsel were on the same side in this situation.


The above-mentioned facts indicate the importance of identifying the sentiment of a statement towards a party in a court case. In the proposed approach, the sentiment of a given phrase is classified into one of the two classes; \textit{negative} and \textit{non-negative}. This classification criterion is selected following the fact that the major use case aligns with classifying terms and entities supporting/referring to either plaintiff or defendant. Therefore, the proposed methodology is focused on identifying the statements with negative sentiment as much as possible. As per requirement, the proposed approach has the ability to extended to explicitly identify the positive sentiment as well following the same methodology. 


In this approach, we propose a novel methodology to perform transfer learning on the RNTN model mentioned in~\citet{socher2013recursive} and build a target model. Given that this is a transfer learning approach, the manually annotated data on movie reviews is used to as the initial source model rather than creating a new comparable manually annotated dataset for the legal domain.


For the testing purposes, we created a manually annotated target domain test dataset such that the phrases belong to one of the two classes: \textit{negative} or \textit{non-negative}. The target system shows a recall of 70.14\% for identifying phrases with negative sentiment in the legal domain. Furthermore, the overall accuracy of the system is above 76\% in classifying sentiments for a given phrase correctly. If this result is compared with the results of source RNTN model~\cite{socher2013recursive}, it is a 6\% improvement in accuracy. The approach proposed in this study can be tried on other domain adaptation tasks related to sentiment classification as well.

\section{Background}
\label{section:background}
Owing to the difficulties in handling legal jargon, efficient and effective computing applications in the field are somewhat sparse. The study by~\citet{schweighofer1993legal} claims that there is a significant vacuum in computerized applications for the field of law which have resulted in an information crisis. The fact that legal vocabulary have words of mixed origin such as English and Latin has been raised as a reason for the difficulty of creating computing applications for the legal domain~\cite{sugathadasa2018legal}.   

However, recently, there have been attempts to involve and build legal ontologies~\cite{jayawardana2017deriving,jayawardana2017semi} as well as attempts to calculate similarity measures in legal domain text~\cite{sugathadasa2017synergistic} and build information retrieval systems thereof~\cite{sugathadasa2018legal}. Given the popularity of knowledge embedding, a number of studies have also attempted to embed legal jargon in vector spaces~\cite{sugathadasa2017synergistic,nay2016gov2vec}. A more recent study by~\citet{ratnayaka2018identifying} uses discourse relations in an attempt to identify relationships among sentences in court case transcripts.

Social media is one of the most used domains for research in sentiment analysis due to the availability of plentiful data. Social media platforms usually contain opinions expressed by people on various topics including politics, sports, entertainment, and others. For instance,~\citet{pak2010twitter} states a research conducted in analyzing language in Twitter posts of millions of users, along with a method to automatically collect a corpus with positive and negative sentiments, where the authors have performed statistical linguistic analysis on the collected corpus and built a sentiment classification system for micro-blogging. They have used a Naive Bayes classifier that uses N-grams and part-of-speech tags as features to train the model. This method is not suitable for analyzing legal text because of the inherent objectivity that needs to be preserved in law.

\textit{Sentiment classification} is also known as \textit{opinion mining}~\cite{esuli2007sentiwordnet}. As such, the study on \textit{Opinion Mining} in legal blogs~\cite{conrad2007opinion} is closest implementation for this study that we have found in the literature. The \textit{Lingpipe} toolkit, of which the sentiment annotation is based on a character-based language model, is used for the sentiment classification in the approach by~\citet{conrad2007opinion}. Further, the data set used for evaluation is based on movie reviews, customer reviews, and MPQA corpus~\cite{wiebe2005annotating}.

SentiWordNet~\cite{esuli2007sentiwordnet,baccianella2010sentiwordnet} classifies synsets of WordNet~\cite{miller1995wordnet} to three classes; \textit{negative}, \textit{positive}, and \textit{objective}. Synsets that do not contain opinionated content are assigned to the \textit{objective} class while the Synsets that do contain opinionated content are assigned to the \textit{subjective} which is then further classified into the two classes \textit{negative} and \textit{positive} depending on the sentiment it carries.  

There have been numerous studies that were built upon SentiWordNet~\cite{esuli2007sentiwordnet,baccianella2010sentiwordnet} which attempts to classify sentiments of phrases and sentences. One such study by \citet{ohana2009sentiment} proposes a methodology to perform opinion mining on movie reviews using support vector machine where some of the features were calculated using WordNet. This achieves an accuracy of 69.35\% and claims that the inaccuracies in SentiWordNet feature calculations are caused by the SentiWordNet's reliance on glosses.~\citet{lu2012unsupervised} evaluates the SentiWordNet for identifying opposing opinion networks in forum discussion. The average SentiWordNet opinion score of words is considered to identify whether a user's expressed comment for a given post has either \textit{for} or \textit{against} relationship. The achieved accuracy using the SentiWordNet opinion score of words is 56\%.

The method proposed by \citet{socher2013recursive} provides an algorithm to identify the sentiment of a phrase or a sentence in a supervised manner using a deep learning model of the type Recursive Neural Tensor Network (RNN). It is claimed that this learning model has the capability to identify the sentiment considering the context of that word. A dataset which consists of movie reviews where each sentence in the data set was broken into phrases and each phrase is annotated by human judges were created for this study. The authors claim a testing accuracy of 80.7\% in phrase level for a test set drawn from the same dataset. Further, the authors claim that the proposed model can be trained over any domain by following the provided methodology. While, theoretically, it is possible, following this for legal domain in a practical implementation which covers a corpus which is both significant and sufficient is difficult. This claim is substantiated by referring the dataset of the original research~\cite{socher2013recursive} which utilized 215,154 manually annotated phrases (from 11,855 sentences) with over 5355 unique words. In comparison to this, the legal corpus used in our study has a vocabulary exceeding 17000 words. The difficulties are not mealy of scale given that the linguistic complexity of legal jargon exceeds that of the average text corpus~\cite{jayawardana2017semi,jayawardana2017word,sugathadasa2017synergistic,sugathadasa2018legal}.   

It is observed that the Recursive Neural Tensor Network (RNTN) model by \citet{socher2013recursive} shows better accuracy in sentiment classification compared to other models. However, the trained model being biased towards the movie reviews which it was trained on is a difficulty that needs to be overcome. For this purpose, several studies~\cite{raina2007self,socher2013zero} claim the process of \textit{domain adaptation} to be a suitable solution. \textit{Domain adaptation} is a sub-category of \textit{Transfer Learning}~\cite{raina2007self}. While the generic process of transfer learning is defined as the process of ``learning model is trained using data from a certain domain and tested with respect to a different domain''~\cite{raina2007self}, the specific case of \textit{domain adaptation} occurs when the Since the task is same in both source and target model. Given that both this study and the original study by \citet{socher2013recursive} works on sentiment classifying, the transfer learning done in this study falls under the definition of domain adaptation~\cite{raina2007self}. Even though transfer learning is not very common in the NLP field, it is extensively used in other fields such as image classification \cite{quattoni2008transfer,raina2007self}.

 The aim of this study is also to build a sentiment classifier specific to the legal domain. But to prepare a manually labeled data set for training purpose is a costly process in terms of time and human effort. Therefore, a \textit{Transfer Learning} approach is used to adapt the RNTN model \cite{socher2013recursive} to the legal domain. When a learning model is trained using data from a certain domain and tested with respect to a different domain, it is called \textit{Transfer Learning} approach~\cite{raina2007self}. Since the task is same in both source~\cite{socher2013recursive} and target model for legal domain, the task belongs to the subcategory called \textit{Domain Adaptation} as mentioned in \cite{raina2007self,socher2013zero}. Image classification \cite{quattoni2008transfer,raina2007self} is a field where transfer learning is vastly used. 

\section{Methodology}
\label{section:methodology}

Given that the transfer learning process described in this study uses the Recursive Neural Tensor Network (RNTN) model proposed by~\citet{socher2013recursive} as the source model, we make numerous references to the aforementioned model throughout the paper. Therefore, to avoid clutter, from this point onward the model proposed by~\citet{socher2013recursive} is referred as \textit{\textbf{Socher Model}} in the remainder of this paper. The main research contributions of this study in the methodological aspect is discussed in this section.

In brief, first, it is required to select the vocabulary from a corpus comprised of legal case transcripts. Then we input a set of words extracted from that corpus to the \textit{Socher Model} for sentiment annotation. After that, three human annotators check for words with deviated sentiments based on the classified classes. Using that identified set, we perform a transfer learning method to identify the sentiment of a given phrase in the legal domain. All these steps are further elaborated in the following sub-sections. 

\subsection{Selecting the Vocabulary}
Depending on the size of the corpus (phrases extracted from legal text), availability of human annotators and the time, it is not feasible to analyze and modify the sentiment of every word in a corpus. Therefore, it is required to select the vocabulary (unique words in the corpus) such that the end-model can correctly classify the sentiment of most of the phrases from the legal domain while not squandering human annotator time on words that occur rarely. To this end, first, the stop-words~\cite{stopwordDefinition} are removed from the text by utilizing the classical stop-word list known as the Van stop-list~\cite{van1979information}. Next, the term frequencies for each word in the corpus is calculated and only the top 95\% words of it are added to the vocabulary. 


\subsection{Assigning Sentiments for the Selected Vocabulary}
\label{subsection:assiging_sentiment_for_the_selected_vocabulary}

The selected vocabulary (set of individual words) is given to the sentiment annotator~\textit{Socher Model} as input. From the model, sentiment is classified into one of the five classes as in table  \ref{table:sentiment-mapping}. This class scheme made sense for the movie reviews for which the~\textit{Socher Model} is trained and used for. However, in the application of this study, the basic requirement of finding sentiment in \textit{court cases} in the legal domain is to identify whether a given statement is against the plaintiff's claim or not. Therefore, we define two classes for sentiment: \textit{negative} and \textit{non-negative}.


Three human judges analyze the selected vocabulary and classify each word into the two classes depending on its sentiment separately and independently. If at least two judges agree, the given word's sentiment is assigned as the class those two judges agreed. For the same word, the output from the sentiment annotator ~\textit{Socher Model} belongs to one of the five classes mentioned in the preceding subsection. In this approach, we map the output from ~\textit{Socher Model} to the two classes we define in Table \ref{table:sentiment-mapping}.

\begin{table}[!htb]
\centering
\label{table:sentiment-mapping}
\begin{tabular}{|l|l|l|}
\hline
& \thead{Human\\annotation} & \thead{\textit{Socher Model}\\output} \\
\hline
\textbf{Class 1} & Negative & \makecell{Very negative,\\negative} \\
\hline
\textbf{Class 2} & Non-negative & \makecell{Neutral,\\Positive,\\very positive} \\
\hline
\end{tabular}
\caption{Sentiment Mapping}
\end{table}

For a given word, if the two sentiment values assigned by the~\textit{Socher Model} and human judges do not agree with the above mapping, we define that the~\textit{Socher Model}'s output has deviated from its actual sentiment. For example: 

\hspace{8mm}\textbf{Sentence:} \textit{Sam is charged with a crime.}

\hspace{8mm}\textbf{\textit{Socher Model}'s output:} positive

\hspace{8mm}\textbf{Human judges' annotation:} negative

The word \textit{charged} has several meanings depending on the context. As the~\textit{Socher Model} was trained using movie reviews, the sentiment of the word \textit{charged} is identified as positive. Although the sentiment of the term \textit{crime} is recognized as negative, the sentiment of the whole sentence is output as positive. But in the legal domain, \textit{charged} refers to a formal accusation. Therefore, the sentiment for the above sentence should have been negative. From the selected vocabulary, all the words with deviated sentiments are identified and listed separately for the further processing.

\subsection{Brief description of the~\textit{Socher Model}}
\label{subsection:description_of_rntn_model}

In the preceding subsection, we came across a situation where the sentiment values from the~\textit{Socher Model} do not match the actual sentiment value because of the difference in domains. And there are words like \textit{insufficient}, which were not recognized by the model because those terms were not included in the training data-set. One approach to solve this is to annotate the phrases extracted from legal case transcripts manually as the ~\textit{Socher Model} suggests, which will require a considerable amount of human effort and time. Instead of that, we can change the model such that the desired output can be obtained using the same trained~\textit{Socher Model} without explicitly training using phrases in the legal domain. Hence, this method is called a transfer learning method.

In order to change the model, first, it is required to understand the internals of the~\textit{Socher Model} model. When a phrase is provided as input, first it generates a binary tree corresponding to the input in which each leaf node represents a single word. Each leaf node is represented as a vector with d-dimensions. The parent nodes are also d-dimensional vectors which are computed in the bottom-up fashion according to some function \textit{g}. The function \textit{g} is composed of a neural tensor layer. Through the training process, the neural tensor layer and the word vectors are adjusted to support the relevant sentiment value. The neural tensor layer corresponds to identify the sentiment according to the structure of words representing the phrase. If we consider a phrase like \textit{not guilty} ,both individual word elements have negative sentiments. But the composition of those words has the structure of negating a negative sentiment term or phrase. Hence the phrase has a non-negative sentiment. If the input was a phrase like \textit{very bad}, the neural tensor layer has the ability to identify that the term \textit{very} increases the negativity in the sentiment. For Example:

\hspace{8mm}\textbf{phrase:} \textit{not guilty.}

\hspace{8mm}\textbf{sentiment:} non-negative

Both words in the above phrase, have negative sentiment if we consider each of them individually. But the composition of those words has the structure of negating a negative sentiment term or phrase. Hence the phrase has a non-negative sentiment. If the input was a phrase like \textit{very bad}, the neural tensor layer has the ability to identify that the term \textit{very} increases the negativity in the sentiment. The hidden process is same as in the preceding example.

\subsection{Adjusting Word Vector Values in RNTN Model}

The requirement of the system is to identify the sentiment of a given phrase. The proposed approach is not to modify the neural tensor layer completely. We simply substitute the word vector values of individual words which are having deviated sentiments between~\textit{Socher Model} and human annotation (See sections \ref{subsection:assiging_sentiment_for_the_selected_vocabulary}). The vectors for the words which were not in the vocabulary of the training set which was used to train the RNTN model should be instantiated. The vectors of the words which are not deviated (according to the definition provided in the preceding subsection \ref{subsection:description_of_rntn_model}) will remain the same.

As the words with deviated sentiments (provided by the~\textit{Socher Model}) in the vocabulary are already known, we initialize the vectors corresponding to the sentiment annotation for those words. Since the model is not trained explicitly, the vector initialization is done by substituting the vectors of words in which sentiment is not deviated comparing the~\textit{Socher Model} output and its actual sentiment. After the substitution is completed, we consider the part-of-speech tag. For that purpose, the part-of-speech tagger mentioned in~\citet{toutanova2003feature} is used. The substitution of vectors is carried out as shown in Table \ref{table:pos_tags_substituted_word_vectors}.

\newcolumntype{Y}{>{\arraybackslash}X}

\begin{table}[!hbt]
\centering
\begin{tabularx}{0.5\textwidth}{|l|Y|Y|}
\hline
\multirow{2}{*}{\thead{POS Tag}}  & \multicolumn{2}{|l|}{\thead{Substituted word vector sentiment}}\\
\hhline{~--}
& \thead{non-negative} & \thead{negative} \\
\hline
JJ & wrong & natural\\ \hline
JJR & worse & natural\\ \hline
JJS & worst & natural\\ \hline
NN & failure & thing \\ \hline
NNS & politics & things \\ \hline
RB & insufficiently & naturally\\ \hline
RBR & insufficiently & naturally\\ \hline
RBS & insufficiently & naturally\\ \hline
VB & hate & do\\ \hline
VBZ & hates & does\\ \hline
VBP & hate & do\\ \hline
VBD & hated & did\\ \hline
VBN & bored & given\\ \hline
VBZ & ignoring & doing\\ \hline
\end{tabularx}
\caption{Substituted Word Vectors for words which should be deviated}
\label{table:pos_tags_substituted_word_vectors}
\break
\end{table}



The number of words which have deviated sentiments is a considerably lower amount compared to the selected vocabulary. The rest of the words' vectors representing sentiments are not changed in the modification process. The neural tensor layer also remains unchanged from the trained~\textit{Socher Model} using movie reviews~\cite{socher2013recursive}. When the vectors for words with deviated sentiments are initialized according to the part-of-speech tag as shown in Table \ref{table:pos_tags_substituted_word_vectors}, it is possible to make a fair assumption that when deciding the sentiment with the proposed implementation, it does not harm the structure corresponding to the linguistic features of English. Consider the sentence ``\textit{evidence is insufficient.}'' as an example. 

\begin{figure}[!ht]
	\centering
	\includegraphics[width=0.4\textwidth]{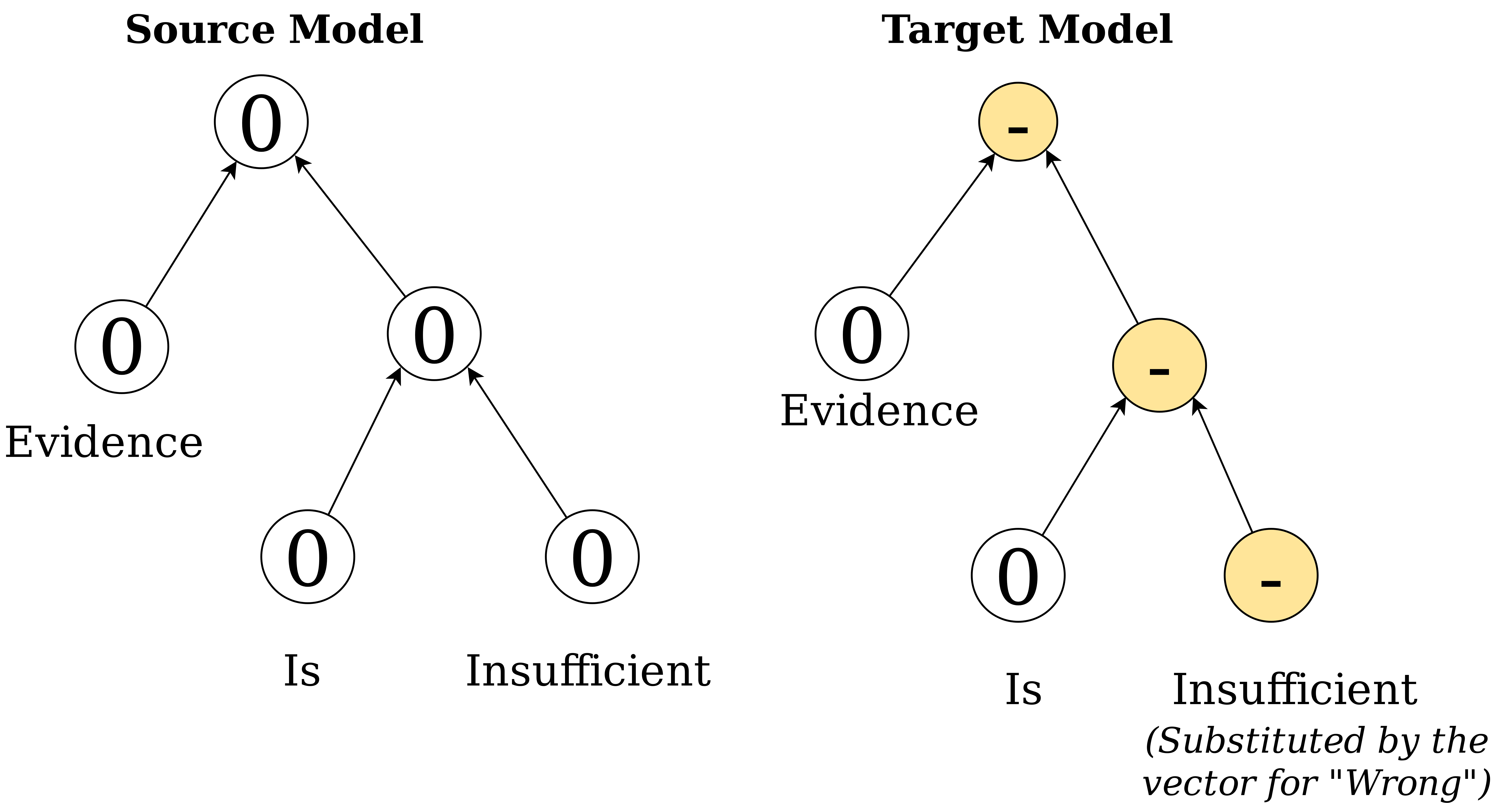}
    \caption{Sentiment Prediction for a phrase with words not in source's vocabulary but in target's vocabulary}
	\label{image:two_graphs_comparison}
	\centering
\end{figure}

The term ``\textit{insufficient}'' is not in the vocabulary of the~\textit{Socher Model} due to the limited vocabulary in training data set. Therefore, the~\textit{Socher Model} provides the sentiment of that word as neutral which indicates as a word with a deviated sentiment. Following the Table \ref{table:pos_tags_substituted_word_vectors}, the sentiment related vector is instantiated by substituting the vector of \textbf{wrong} as the part-of-speech tag of \textbf{insufficient} is \textbf{JJ}~\cite{santorini1990part}. Therefore the modified version of the RNTN model has the capability of identifying the sentiment of the above sentence as negative. The figure \ref{image:two_graphs_comparison} shows how the sentiment is induced through the newly instantiated word vector. 

And there are scenarios where the term is in the vocabulary of the~\textit{Socher Model} but has a different sentiment compared to the legal domain. Consider the sentence ``\textit{Sam is charged with a crime}'' which was mentioned in section \ref{subsection:assiging_sentiment_for_the_selected_vocabulary}.

\begin{figure}[!ht]
	\centering
	\includegraphics[width=0.4\textwidth]{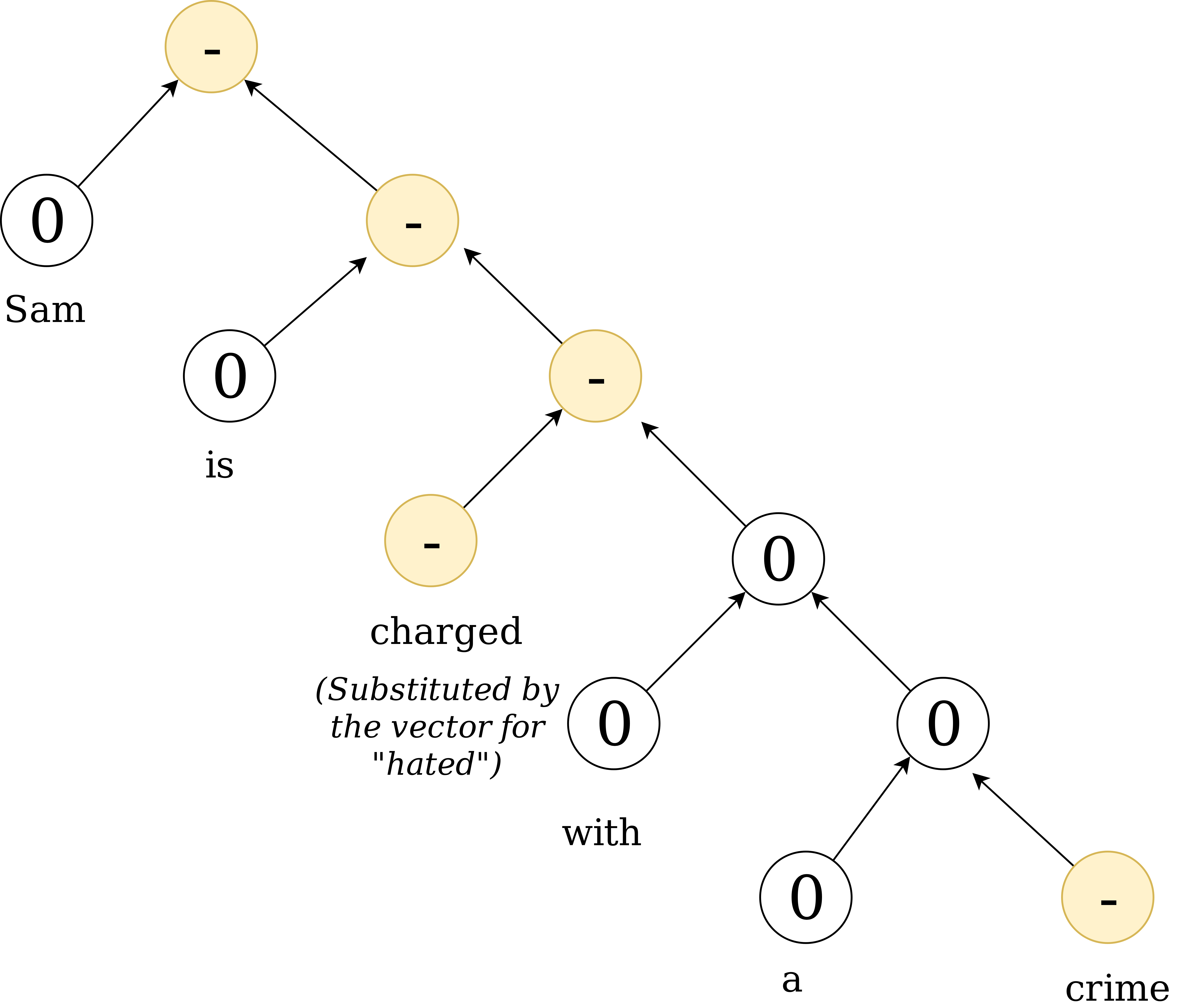}
    \caption{Sentiment Prediction for a phrase with words having deviated sentiment in two domains - target model \label{image:improved-model}}
	\centering
\end{figure}

In section \ref{subsection:assiging_sentiment_for_the_selected_vocabulary}, we have identified that the term \textit{charged} denotes a different sentiment in legal domain compared to movie reviews. The source RNTN model outputs a positive sentiment for that given sentence as the term \textit{charged} is identified as having a positive sentiment according to movie reviews domain. And that term is the cause for having such an output from the source model. The figure \ref{image:improved-model} indicates how the change we introduced in the target model (in section \ref{subsection:assiging_sentiment_for_the_selected_vocabulary}) induce the correct sentiment up to the root level of the phrase. Therefore, the target model identifies the sentiment correctly for the given phrase. 

\begin{figure}[!ht]
	\centering
	\includegraphics[width=0.4\textwidth]{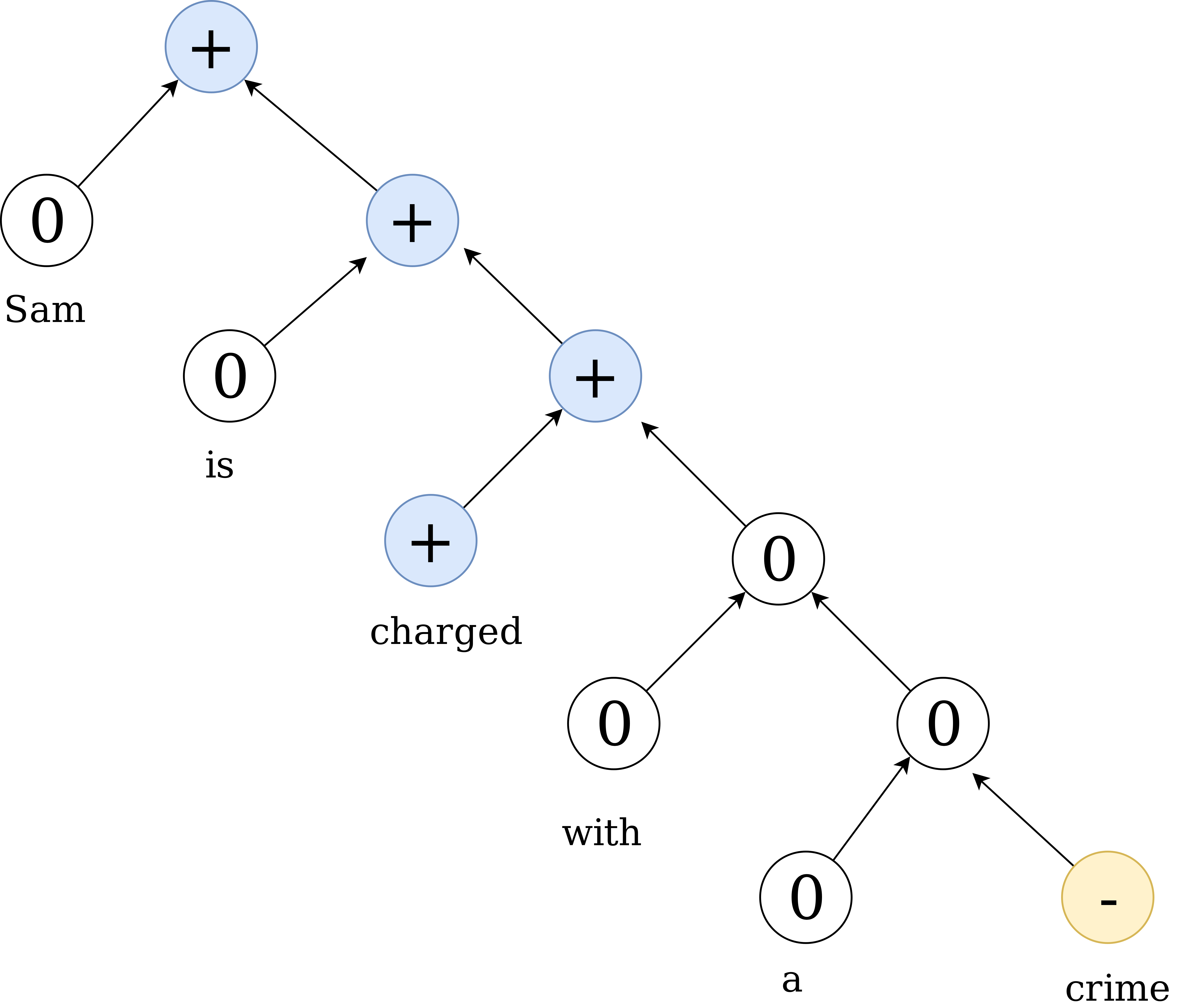}
    \caption{Sentiment Prediction for a phrase with words having deviated sentiment in two domains - source model \label{image:original-model}}
	
	\centering
\end{figure}

To improve the recall in identifying phrases with negative sentiment, we have added another rule to the classification criteria. The source RNTN model (\textit{Socher Model}) provides the score for each of the five classes such that all those five scores sum up to 1. If the negative sentiment class has the highest score, the sentiment label of the phrase will be \textit{negative}. Otherwise, the phrase again can be classified as having a \textit{negative} sentiment if the score for negative sentiment class is above 0.4. If those two conditions are not met, the phrase will be classified as having a \textit{non-negative} sentiment. Section \ref{section:experiments-and-results} provides observations and results regarding the improved criteria.

\section{Experiments and Results}
\label{section:experiments-and-results}
The proposed approach in this paper is based on transfer learning. Therefore, we needed to create a golden standard for identifying sentiments of phrases and sentences in the legal domain in order to evaluate the model. The phrases and sentences for the test data set are randomly picked from legal case transcripts based on the United States Supreme Court. During the selection process, we have selected an equal amount of phrases for both classes according to the~\textit{Socher Model}. Each of these phrases and sentences is annotated by three human annotators. Since the classification process is binary, we pick the sentiment class for each test subject based on the maximum number of votes. In the end, we prepare the test data set containing nearly 1500 annotations to use in the evaluation process. 

In the experiment, we compare the sentiment class picked by human judges and the modified RNTN model. As the baseline model, we use the source RNTN model (\textit{Socher Model}) to check the impact caused by the proposed transfer learning approach. The acquired results from the baseline model is shown in Table \ref{table:confusion_matrix_original_model} and results from the target model is shown in Table \ref{table:confusion_matrix_improved_model}.

According to Table \ref{table:confusion_matrix_original_model} and Table \ref{table:confusion_matrix_improved_model}, there is a 10\% improvement in identifying phrases with negative sentiment. The reason is that there are a lot of unknown words which are in the legal domain but not in movie reviews corpus. In addition, we have introduced new criteria based on a threshold for the score of negative class to improve the recall. Due to that reason, the precision in identifying phrases with a negative sentiment is 84.41\%. But if we compare with the precision of the baseline model (\textit{Socher Model}) for negative sentiment class is 79.62\% which is a lower value. Since the test dataset is not skewed a lot towards one class, it is fair to consider the accuracy of the system in predicting the sentiment for any given phrase. The baseline model shows the accuracy of 70.17\% while the target model shows 76.80\%. The improvement in accuracy is above 6\%.

\newcommand{\shadedCell}[1]{
#1\% 
\cellcolor{gray!#1!white}
}

\begin{table}[h]
\begin{center}
\begin{tabular}{|c|c|c|c|}
\hline
\diagbox[width=7em]{Actual}{Predicted}
 & \thead{Negative} & \thead{Non\\negative} & \thead{Total}  \\ 
 \hline
\textbf{Negative} & \shadedCell{60.43} & \shadedCell{39.57} & 278  \\ \cline{1-4}
\textbf{Non-negative} & \shadedCell{18.29} & \shadedCell{81.71} & 235  \\ \cline{1-4}
\thead{Total} & 211 & 301 & 513 \\ \hline
\end{tabular}
\caption{Confusion Matrix for Results from the Baseline Model}
\label{table:confusion_matrix_original_model}
\end{center}
\end{table}

\begin{table}[h]
\begin{center}
\begin{tabular}{|c|c|c|c|}
\hline
\diagbox[width=7em]{Actual}{Predicted}
 & \thead{Negative} & \thead{Non\\negative} & \thead{Total}  \\ 
 \hline
\textbf{Negative} & \shadedCell{70.14} & \shadedCell{29.86} & 278  \\ \cline{1-4}
\textbf{Non-negative} & \shadedCell{15.32} & \shadedCell{84.68} & 235  \\ \cline{1-4}
\thead{Total} & 231 & 282 & 513 \\ \hline
\end{tabular}
\caption{Confusion Matrix for Results from the Improved Model}
\label{table:confusion_matrix_improved_model}
\end{center}
\end{table}

The observed results in Table \ref{table:confusion_matrix_original_model} and Table \ref{table:confusion_matrix_improved_model} show that there is a 6\% improvement of the sentiment with respect to the baseline model. There are a few reasons behind the results. As we randomly selected phrases from the legal case transcripts corpus, only 45\% of the phrases actually contained the words where we had substituted the vector regarding sentiment. Therefore, the output for 55\% of the phrases from the baseline model and the target model was the same. If we compare the output provided by the baseline model and the target model, output of 9.5\% of the total phrases are different to each other. Therefore the difference between the two models is based on that 9.5\% of the total phrases.

\section{Conclusion and Future Work}
\label{section:conclusions}
This study is focused on building an automatic sentiment annotator for legal texts based on the \textit{Recursive Neural Tensor Network (RNTN)} model mentioned in~\citet{socher2013recursive}. Furthermore, this study can be identified as a transfer learning approach as it is not required to prepare a training data set for the legal domain specifically. Instead, this approach uses the same training data set stated in~\citet{socher2013recursive}. This task can be recognized as a domain adaptation task. The proposed approach could achieve a 70.14\% recall in identifying phrases with negative sentiments (improvement is 10\% compared to the source model). The accuracy of the target model is above 76\% which is a 6\% improvement over the source model. 

The proposed methodology can be adjusted for any domain adaptation task other than the legal domain, which makes this study more important. To train the model, it is not required to prepare manually annotated data for a specific domain. Another advantage is that if there are improvements introduced to the source model, those improvements can be inherited to the target model as well. The major disadvantage associated with this model is that the accuracy of the target model will be limited by the source model in most occasions. In other words, it is hard to exceed the accuracy shown by the source model for its own domain.

There are words which produce one sentiment when they are combined but provide completely different sentiments when considered as individual elements. If we consider the term ``cover up'' in the legal domain, it has a meaning of hiding some mistake or crime. Therefore, it should have a negative sentiment. But the individual terms do not indicate negative sentiment. Therefore, the results can be further improved by considering bi-grams and tri-grams.

The improved version of the Stanford CoreNLP sentiment annotator \cite{socher2013recursive} could be used for further research on using machine learning for the legal domain. Furthermore, the transfer learning method we have described in this study is adjustable for any domain to build an automated sentiment annotator.


\bibliography{references}
\bibliographystyle{acl_natbib_nourl}

\end{document}